\documentclass[conference,a4paper]{IEEEtran}

\usepackage[
pdftitle={Biology of Applied Digital Ecosystems}, 
pdfauthor={Gerard Briscoe},
pdfsubject={Ecosystem-Oriented Architectures},
colorlinks=true, linkcolor=black, citecolor=black,  filecolor=black, urlcolor=black,
hyperfootnotes=false]{hyperref}

\usepackage{graphicx, setspace, acronym, ifthen, substr, multibib, color, amsmath, tocloft}

\newcommand{\execute}[1]{\immediate\write18{#1}}

\newboolean{final}\setboolean{final}{true}

\newcommand{\tfigure}[9]
	{
	\IfSubStringInString{!}{#8}{\begin{figure}[#8]\IfSubStringInString{mm}{#9}{\vspace{#9}}{}}{\begin{figure}[!t]}
	\centering
	
	\IfSubStringInString{pdf}{#3}
		{
		\execute{cd images; ln -s #2.pdf .#2.gdf}
		\href{file://localhost/Users/g/Desktop/agentStability/images/.#2.gdf}{\includegraphics[#1]{images/#2}}
		}
		{\IfSubStringInString{graph}{#3}
			{
			\execute{cd images; ./makeGraph.sh #2; ln -s #2.pdf .#2.gdf}
			\ifthenelse{\boolean{final}}
				{\includegraphics[#1]{images/#2}}
				{\href{file://localhost/Users/g/Desktop/agentStability/images/.#2.gdf}{\includegraphics[#1]{images/#2}}}
			}
			{
			\execute{cd images; ./pdfcrop.sh #2}
			\ifthenelse{\boolean{final}}
				{\includegraphics[#1]{images/#2-crop.pdf}}
				{\href{file://localhost/Users/g/Desktop/agentStability/images/#2.#3}{\includegraphics[#1]{images/#2-crop}}}
			}
		}
		
	\vspace{#7}
	\caption[#4]
		{
		\label{#6}
		\tcaption{#4}{#5}
		}
	\IfSubStringInString{!t}{#8}{\IfSubStringInString{mm}{#9}{\vspace{#9}}{}}{}
	\end{figure}
	}

\newcommand{\italics}{\textit}

\newcommand{\tcaption}[2]
	{
	\IfSubStringInString{:}{#2}{\italics{#1 #2}}{\italics{#1: #2}}
	}

\newcommand{\setCap}[2]{#1\immediate\write18{./mkcaption.sh #2}}
\newcommand{\getCap}[1]{\acl{#1}}

\begin{document}

\acrodef{ABM}{Agent-Based Model}
\acrodef{AI}{Artificial Intelligence}
\acrodef{DAI}{Distributed Artificial Intelligence}
\acrodef{API}{Application Programming Interface}
\acrodef{ARF}{p14ARF human tumor-suppressor gene}
\acrodef{B2B}{business-to-business}
\acrodef{BDP}{Biologically inspired Design Pattern}
\acrodef{BGS}{Best Guess Solution}
\acrodef{BIC}{Biologically-Inspired Computing}
\acrodef{BML}{Business Modelling Language}
\acrodef{BPEL}{Business Process Execution Language}
\acrodef{BPMN}{Business Process Modeling Notation}
\acrodef{CAS}{Complex Adaptive Systems}
\acrodef{COBOL}{COmmon Business-Oriented Language}
\acrodef{DBE}{Digital Business Ecosystem}
\acrodef{DE}{Digital Ecosystem}
\acrodef{DEC}{distributed evolutionary computing}
\acrodef{DGA}{Distributed genetic algorithms}
\acrodef{DIS}{Distributed Intelligence System}
\acrodef{DNA}{Deoxyribose Nucleic Acid}
\acrodef{DOP}{DBE Open Protocol}
\acrodef{DSS}{Distributed Storage System}
\acrodef{EAP}{Evolving Agent Population}
\acrodef{ebXML}{e-business eXtensible Markup Language}
\acrodef{EC}{Evolutionary Computing}
\acrodef{ECJ}{Evolutionary Computing in Java}
\acrodef{EE}{Evolutionary Environment}
\acrodef{EFL}{Evolutionary Framework for Language}
\acrodef{FLE}{Framework for Language Ecosystems}
\acrodef{EOA}{Ecosystem-Oriented Architecture}
\acrodef{ESS}{evolutionary stable strategy}
\acrodef{EvE}{Evolutionary Environment}
\acrodef{ExE}{Execution Environment}
\acrodef{FCB}{Framework for Computational Biomimicry}
\acrodef{FFF}{Fitness Function Framework}
\acrodef{FL}{Fitness Landscape}
\acrodef{HWU}{Heriot-Watt University}
\acrodef{ICL}{Imperial College London}
\acrodef{ICT}{Information and Communications Technology}
\acrodef{INTEL}{Intel Ireland}
\acrodef{IPA}{International Phonetic Alphabet}
\acrodef{ISUFI}{Istituto Superiore Universitario di Formazione Interdisciplinare}
\acrodef{JDJ}{Java Developer's Journal}
\acrodef{KC}{Kolmogorov-Chaitin}
\acrodef{LAN}{local area network}
\acrodef{LSE}{London School of Economics and Political Science}
\acrodef{MAS}{Multi-Agent System}
\acrodef{MDL}{Minimum Description Length}
\acrodef{MDM2}{murine double minute 2}
\acrodef{MFT}{Mean Field Theory}
\acrodef{MoAS}{Mobile Agent System}
\acrodef{MOF}{Meta Object Facility}
\acrodef{MUH}{migration and usage history}
\acrodef{NIC}{Nature Inspired Computing}
\acrodef{NN}{Neural Network}
\acrodef{NoE}{Network of Excellence}
\acrodef{OMG}{Open Mac Grid}
\acrodef{OPAALS}{Open Philosophies for Associative Autopoietic Digital Ecosystems}
\acrodef{P2P}{peer-to-peer}
\acrodef{P53}{protein 53}
\acrodef{PDA}{Personal Digital Assistant}
\acrodef{QoS}{quality of service}
\acrodef{REST}{REpresentational State Transfer}
\acrodef{RNA}{Deoxyribose Nucleic Acid}
\acrodef{SAE}{Software Agent Ecosystem}
\acrodef{SBML}{Systems Biology Modelling Language}
\acrodef{SBVR}{Semantics of Business Vocabulary and Business Rules}
\acrodef{SDL}{Service Description Language}
\acrodef{SF}{Service Factory}
\acrodef{SIM}{Social Interaction Mechanism}
\acrodef{SM}{Service Manifest}
\acrodef{SME}{Small and Medium sized Enterprise}
\acrodef{SML}{Service Modelling Language}
\acrodef{SMO}{Sequential Minimal Optimisation}
\acrodef{SOA}{Service-Oriented Architecture}
\acrodef{SOAP}{Simple Object Access Protocol}
\acrodef{SOC}{Self-Organised Criticality}
\acrodef{SOLUTA}{SOLUTA.NET}
\acrodef{SOM}{Self-Organising Map}
\acrodef{SSL}{Semantic Service Language}
\acrodef{STU}{Salzburg Technical University}
\acrodef{SUN}{Sun Microsystems}
\acrodef{SVM}{Support Vector Machines}
\acrodef{TM}{Turing Machine}
\acrodef{UBHAM}{University of Birmingham}
\acrodef{UDDI}{Universal Description Discovery and Integration}
\acrodef{UML}{Unified Modelling Language}
\acrodef{URI}{Uniform Resource Identifier}
\acrodef{UTM}{Universal Turing Machine}
\acrodef{VLP}{variable length population}
\acrodef{VLS}{variable length sequences}
\acrodef{vls}{variable length sequence}
\acrodef{WP}{Work-Package}
\acrodef{WSDL}{Web Services Definition Language}
\acrodef{XMI}{XML Metadata Interchange}
\acrodef{XML}{eXtensible Markup Language}
\acrodef{MD5}{Message-Digest algorithm 5}
\acrodef{GA}{genetic algorithm}
\acrodef{GP}{genetic programming}
\acrodef{MASON}{Multi-Agent Simulator Of Neighborhoods}
\acrodef{Repast}{Recursive Porous Agent Simulation Toolkit}
\acrodef{JCLEC}{Java Computing Library for Evolutionary Computing}
\acrodef{OWL-S}{Web Ontology Language - Service}
\acrodef{EGT}{Evolutionary Game Theory}

\title{Biology of Applied Digital Ecosystems\pdfbookmark[1]{Title}{Title}}

\author{
\authorblockN{Gerard Briscoe}
\authorblockA{Intelligent Systems and Networks Group\\
Department of Electrical and Electronic Engineering\\
Imperial College London\\
London, United Kingdom\\ 
e-mail: gerard.briscoe@ic.ac.uk}
\and 
\authorblockN{Suzanne Sadedin, Greg Paperin}
\authorblockA{Clayton School of Information Technology\\
Monash University\\
Melbourne, Victoria 3800, Australia\\
e-mail: suzanne.sadedin@infotech.monash.edu.au,\\ gregory.paperin@infotech.monash.edu.au}
}
\maketitle

\begin{abstract}
\pdfbookmark[1]{Abstract}{Abstract}
A primary motivation for our research in Digital Ecosystems is the desire to exploit the self-organising properties of biological ecosystems. Ecosystems are thought to be robust, scalable architectures that can automatically solve complex, dynamic problems. However, the biological processes that contribute to these properties have not been made explicit in Digital Ecosystems research. Here, we discuss how biological properties contribute to the self-organising features of biological ecosystems, including population dynamics, evolution, a complex dynamic environment, and spatial distributions for generating local interactions. The potential for exploiting these properties in artificial systems is then considered. We suggest that several key features of biological ecosystems have not been fully explored in existing digital ecosystems, and discuss how mimicking these features may assist in developing robust, scalable self-organising architectures. An example architecture, the Digital Ecosystem, is considered in detail. The Digital Ecosystem is then measured experimentally through simulations, with measures originating from theoretical ecology, to confirm its likeness to a biological ecosystem. Including the responsiveness to requests for applications from the user base, as a measure of the \emph{ecological succession} (development).

\emph{Index Terms--} evolution, ecosystem, complexity

\end{abstract}
\IEEEpeerreviewmaketitle

\acrodef{archComTop}{many strongly connected clusters (communities), called {sub-networks} (quasi-complete graphs), with a few connections between these clusters (communities) \cite{swn1}. Graphs with this topology have a very high clustering coefficient and small characteristic path lengths \cite{swn1}.}
\acrodef{similarCap}{requests are evaluated on separate {islands} (populations), and so adaptation is accelerated by the sharing of solutions between evolving populations (islands), because they are working to solve similar requests (problems).}
\acrodef{picUser}{will formulate queries to the Digital Ecosystem by creating a request as a {semantic description}, like those being used and developed in \acp{SOA} \cite{SOAsemantic}, specifying an application they desire and submitting it to their local peer (habitat).}
\acrodef{picUserReq}{A population is then instantiated in the user's habitat in response to the user's request, seeded from the agents available at their habitat.}
\acrodef{semanticRequest}{A simulated user request consisted of an abstract {semantic description}, as a list of sets of numeric tuples to represent the properties of a desired business application}
\acrodef{succession}{So, it becomes increasingly more complex through this process of succession, driven by the evolution of the populations within the ecosystem \cite{connell111msn}.}
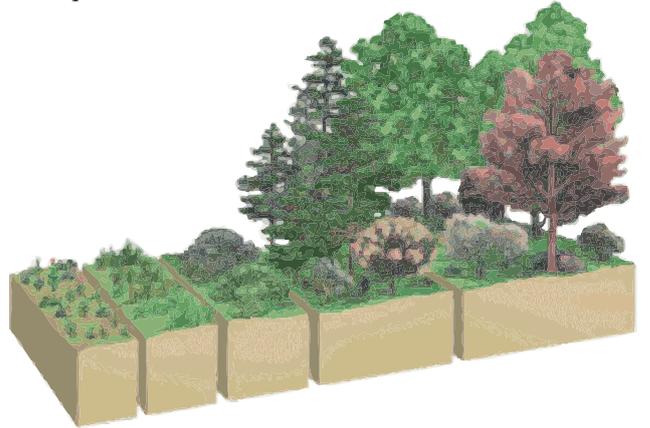
\acrodef{succession2}{The formation of a mature ecosystem}
\acrodef{succession3}{is the slow, predictable, and orderly changes in the composition and structure of an ecological community, for which there are defined stages in the increasing complexity \cite{begon96}, as shown}
\acrodef{DigEcoSuc2}{the end of the simulation run, the agent-sequences had evolved and migrated over only ten generations, on average, and collectively had already reached near 70\% effectiveness for the user requests.}
\acrodef{DigEcoSuc}{The formation of a mature biological ecosystem, ecological succession, is a relatively slow process \cite{begon96}, and the simulated Digital Ecosystem acted similarly in reaching a mature state.}
\acrodef{speciesAbundance}{is the proportion of all organisms, in a community, belonging to a particular species \cite{Bell}. Relative abundance distributions provide a measure of inequalities in population size within an ecosystem, and in most biological ecosystems this distribution takes a log-normal form \cite{Bell}.}
\acrodef{specAbund2}{the Digital Ecosystem did not conform to the expected log-normal}
\acrodef{speciesArea}{The {species-area} relationship measures diversity relative to spatial scale \cite{sizling2004pls}. In the Digital Ecosystem, this relationship represents how similar solutions are to one another at different habitat scales. The species-area relationship is commonly found to follow a power law in biological ecosystems}
\acrodef{ecoClass}{then the {Digital Ecosystem} and {biological ecosystem} classes would both inherit from the abstract {ecosystem} class, but implement its attributes differently}
\acrodef{eco2Class}{So, we would argue that the apparent compromises in mimicking biological ecosystems are actually features unique to Digital Ecosystems.}

\immediate\write18{echo > captions.tex}

\section{Introduction}
Is mimicking ecosystems the future of information systems? A key challenge in modern computing is to develop systems that address complex, dynamic problems in a scalable and efficient way, because the increasing complexity of software makes designing and maintaining efficient and flexible systems a growing challenge \cite{newsArticle1, slashdot, newsArticle3}. What with the ever expanding number of services being offered online from \acp{API} being made public, there is an ever growing number of computational units available to be combined in the creation of applications. However, this is currently a task done manually by programmers, and it has been argued that current software development techniques have hit a \emph{complexity wall} \cite{lyytinen2001nwn}, which can only be overcome by automating the search for new algorithms. There are several existing efforts aimed at achieving this automated service composition \cite{reef3, reef5, reef1, reef4}, the most prevalent of which is \aclp{SOA} and its associated standards and technologies \cite{curbera2002uws, SOAstandards}. 

Alternatively, nature has been in the research business for 3.8 billion years and in that time has accumulated close to 30 million \emph{well-adjusted} solutions to a plethora of design challenges that humankind struggles to address with mixed results \cite{biomimicry}. Biomimicry is a discipline that seeks solutions by emulating nature's designs and processes, and there is considerable opportunity to learn elegant solutions for human-made problems \cite{biomimicry}. Biological ecosystems are thought to be robust, scalable architectures that can automatically solve complex, dynamic problems, possessing several properties that may be useful in automated systems. These properties include self-organisation, self-management, scalability, the ability to provide complex solutions, and automated composition of these complex solutions \cite{Levin}.

Therefore, an approach to the aforementioned challenge would be to develop Digital Ecosystems, artificial systems that aim to harness the dynamics that underlie the complex and diverse adaptations of living organisms in biological ecosystems. While evolution may be well understood in computer science under the auspices of \emph{evolutionary computing} \cite{eiben2003iec}, ecological models are not. The possible connections between Digital Ecosystems and their biological counterparts are yet to be closely examined, so potential exists to create an \acl{EOA} with the essential elements of biological ecosystems, where the word \emph{ecosystem} is more than just a \emph{metaphor}. We propose that an ecosystem inspired approach, would be more effective at greater scales than traditionally inspired approaches, because it would be built upon the scalable and self-organising properties of biological ecosystems \cite{Levin}.

Our focus is in creating the digital counterpart of biological ecosystems. However, the term \emph{digital ecosystem} has been used to describe a variety of concepts, which it now makes sense to review. Some of these refer to the existing networking infrastructure of the internet \cite{debook2, fiorina, XIMBIOTIX}, while several companies offer a \emph{digital ecosystem} service or solution, which involves enabling customers to use existing e-business solutions \cite{accenture, syntel, xewow}. The term is also being increasingly linked, yet undefined, to the future developments of \ac{ICT} adoption for e-business and e-commerce, to create so called \emph{business ecosystems} \cite{iansiti2004kan, nachira, papazoglou2001aot}. However, perhaps the most frequent references to \emph{digital ecosystems} arise in Artificial Life research, where they are created primarily to investigate aspects of biological and other complex systems \cite{sorakugun1995eas, grand1998ces, deAI1}. 

The extent to which these disparate systems resemble biological ecosystems varies, and frequently the word \emph{ecosystem} is merely used for branding purposes without any inherent ecological properties. We consider Digital Ecosystems to be software systems that exploit the properties of biological ecosystems, and suggest that several key features of biological ecosystems have not been fully explored in existing \emph{digital ecosystems}. So, we will now discuss how mimicking these features can create Digital Ecosystems, which are robust, scalable, and self-organising.

Arguably the most fundamental differences between biological and digital ecosystems lie in the motivation and approach of their respective researchers. Biological ecosystems are ubiquitous natural phenomena whose maintenance is crucial to our survival, developing through the process of \emph{ecological succession} \cite{begon96}. In contrast, Digital Ecosystems will be defined here as a technology engineered to serve specific human purposes, developing to solve dynamic problems in parallel with high efficiency.

Genetic algorithms are a form of evolutionary computing, and like all forms uses natural selection to evolve solutions \cite{goldberg}; started with a set of possible solutions chosen arbitrarily, then selection, replication, recombination, and mutation are applied iteratively. Selection is based on conforming to a fitness function which is determined by a specific problem of interest, and so over time better solutions to the problem can thus evolve \cite{goldberg}. As Digital Ecosystems will likely solve problems by evolving solutions, they will probably incorporate some form of evolutionary computing. However, we suggest that Digital Ecosystems should also incorporate additional features, providing it with a closer resemblance to biological ecosystems. Including features such as complex dynamic fitness functions, a distributed or network environment, and self-organisation arising from interactions among organisms and their environment, which we will discuss later.

\section{Fitness Landscapes and Agents}

\label{agents}
An ecosystem comprises both an environment and a set of interacting, reproducing entities (or agents) in that environment; with the environment acting as a set of physical and chemical constraints on reproduction and survival \cite{begon96}. These constraints can be considered in abstract using the metaphor of the fitness landscape, in which individuals are represented as solutions to the problem of survival and reproduction \cite{wright1932}. All possible solutions are distributed in a space whose dimensions are the possible properties of individuals. An additional dimension, height, indicates the relative fitness (in terms of survival and reproduction) of each solution. The fitness landscape is envisaged as a rugged, multidimensional landscape of hills, mountains, and valleys, because individuals with certain sets of properties are \emph{fitter} than others \cite{wright1932}. 

In biological ecosystems, fitness landscapes are virtually impossible to identify. This is both because there are large numbers of possible traits that can influence individual fitness, and because the environment changes over time and space \cite{begon96}. In contrast, within a digital environment, it is normally possible to specify explicitly the constraints that act on individuals in order to evolve solutions that perform better within these constraints. Within genetic algorithms, exact specification of a fitness landscape or function is common practice \cite{goldberg}. However, within a Digital Ecosystem the ideal constraints are those that allow solution populations to evolve to meet user needs with maximum efficiency. User needs will change from place to place and time to time. In this sense the fitness landscape of a Digital Ecosystem is complex and dynamic, and more like that of a biological ecosystem than like that of a traditional genetic algorithm \cite{morrison2004dea, goldberg}. The designer of a Digital Ecosystem therefore faces a double challenge: firstly, to specify rules that govern the shape of the fitness function/landscape in a way that meaningfully maps landscape dynamics to user requests, and secondly, to evolve within this space, solution populations that are diverse enough to solve disparate problems, complex enough to meet user needs, and efficient enough to be preferable to those generated by other means.

The agents within a Digital Ecosystem will need to be like biological individuals in the sense that they reproduce, vary, interact, move, and die \cite{begon96}. Each of these properties contributes to the dynamics of the ecosystem. However, the way in which these individual properties are encoded may vary substantially depending on the intended purpose of the system \cite{chmbers2001phg}.

\section{Networks and Spatial Dynamics}

A key factor in the maintenance of diversity in biological ecosystems is spatial interactions, and several modelling systems have been used to represent these spatial interactions. Including metapopulations\footnote{A metapopulation is a collection of relatively isolated, spatially distributed, local populations bound together by occasional dispersal between populations. \cite{levins1969sda, hanski1999me, hanski2003mtf}}, diffusion models, cellular automata and agent-based models (termed individual-based models in ecology) \cite{Greenetal2006}. The broad predictions of these diverse models are in good agreement. At local scales, spatial interactions favor relatively abundant species disproportionately. However, at a wider scale, this effect can preserve diversity, because different species will be locally abundant in different places. The result is that even in homogeneous environments, population distributions tend to form discrete; long-lasting patches that can resist an invasion by superior competitors \cite{Greenetal2006}. Population distributions can also be influenced by environmental variations such as barriers, gradients, and patches. The possible behaviour of spatially distributed ecosystems is so diverse that scenario-specific modelling is necessary to understand any real system \cite{suzie}. Nonetheless, certain robust patterns are observed. These include the relative abundance of species, which consistently follows a roughly log-normal relationship  \cite{Bell}, and the relationship between geographic area and the number of species present, which follows a power law \cite{sizling2004pls}. The reasons for these patterns are disputed, because they can be generated by both spatial extensions of simple Lotka-Volterra competition models \cite{Hubbell}, and more complex ecosystem models \cite{Sole}. 

Landscape connectivity plays an important part in ecosystems. When the density of habitats within an environment falls below a critical threshold, widespread species may fragment into isolated populations. Fragmentation can have several consequences. Within populations, these effects include loss of genetic diversity and detrimental inbreeding \cite{GreenKirley}. At a broader scale, isolated populations may diverge genetically, leading to speciation. 

From an information theory perspective, this phase change in landscape connectivity can mediate global and local search strategies \cite{Greenetal2000}. In a well-connected landscape, selection favors the globally superior, and pursuit of different evolutionary paths is discouraged, potentially leading to premature convergence. When the landscape is fragmented, populations may diverge, solving the same problems in different ways. Recently, it has been suggested that the evolution of complexity in nature involves repeated landscape phase changes, allowing selection to alternate between local and global search \cite{Greenetalinpress}. 

In a digital context, we can have spatial interactions by using a distributed system that consists of a set of interconnected locations, with agents that can migrate between these connected locations. In such systems the spatial dynamics are relatively simple compared with those seen in real ecosystems, which incorporate barriers, gradients, and patchy environments at multiple scales in continuous space \cite{begon96}. Nevertheless, depending on how the connections between locations are organised, such Digital Ecosystems might have dynamics closely parallel to spatially explicit models, diffusion models, or metapopulations \cite{suzie}. We will discuss later the use of a dynamic non-geometric spatial network, and the reasons for using this approach.

\section{Selection and Self-Organization}

The major hypothetical advantage of Digital Ecosystems over other complex organisational models is their potential for dynamic adaptive self-organisation. However, for the solutions evolving in Digital Ecosystems to be useful, they must not only be efficient in a computational sense, but they must also solve purposeful problems. That is, the fitness of agents must translate in some sense to real-world usefulness as demanded by the users \cite{ducheyne2003fiu}.

Constructing a useful Digital Ecosystem therefore requires a balance between freedom of the system to self-organise, and constraint of the system to generate useful solutions. These factors must be balanced because the more the system's behaviour is dictated by its internal dynamics, the less it may respond to fitness criteria imposed by the users. At one extreme, when system dynamics are mainly internal, agents may evolve that are good at survival and reproduction within the digital environment, but useless in the real world \cite{ducheyne2003fiu}. At the other extreme, where the users' fitness criteria overwhelmingly dictates function, we suggest that dynamic exploration, of the solution space and complexity, is likely to be limited. The reasoning behind this argument is as follows. Consider a multidimensional solution space which maps to a rugged fitness landscape \cite{wright1932} . In this landscape, competing solution lineages will gradually become extinct through chance processes. So, the solution space explored becomes smaller over time as the population adapts and the diversity of solutions decreases. Ultimately, all solutions may be confined to a small region of the solution space. In a static fitness landscape, this situation is not undesirable because the surviving solution lineages will usually be clustered around an optimum \cite{goldberg}. However, if the fitness landscape is dynamic, the location of optima varies over time, and should lineages become confined to a small area of the solution space, then subsequent selection will locate only optima that are near this area \cite{morrison2004dea}. This is undesirable if new, higher optima arise that are far from pre-existing ones. A related issue is that complex solutions are less likely to be found by chance than simple ones. Complex solutions can be visualised as sharp, isolated peaks on the fitness landscape. Especially for dynamic landscapes, these peaks are most likely to be found when the system explores the solution space widely \cite{morrison2004dea}. Therefore, a self-organising mechanism other than the fitness criteria of users is required to maintain diversity among competing solutions in a Digital Ecosystem.

\section{Complex Adaptive Systems: Stability/Diversity}

Ecosystems are often described as \ac{CAS}, because like them, they are systems made from diverse, locally interacting components that are subject to selection. Other \ac{CAS} include brains, individuals, economies, and the biosphere. All are characterised by hierarchical organisation, continual adaptation and novelty, and non-equilibrium dynamics. These properties lead to behaviour that is non-linear, historically contingent, subject to thresholds, and contains multiple basins of attraction \cite{Levin}. 

In the previous subsections, we have advocated Digital Ecosystems that include agent populations evolving by natural selection in distributed environments. Like real ecosystems, digital systems designed in this way fit the definition of \ac{CAS}. The features of these systems, especially non-linearity and non-equilibrium dynamics, offer both advantages and hazards for adaptive problem-solving. The major hazard is that the dynamics of \ac{CAS} are intrinsically hard to predict because of the non-linear emergent self-organisation \cite{levin1999fdc}. This observation implies that designing a useful Digital Ecosystem will be partly a matter of trial and error. The occurrence of multiple basins of attraction in \acp{CAS} suggests that even a system that functions well for a long period may suddenly at some point transition to a less desirable state \cite{folke}. For example, in some types of system self-organising mass extinctions might result from interactions among populations, leading to temporary unavailability of diverse solutions \cite{newman1997mme}. This concern may be addressed by incorporating negative feedback or other mechanisms at the global scale. The challenges in designing an effective Digital Ecosystem are mirrored by the system's potential strengths. Non-linear behaviour provides the opportunity for scalable organisation and the evolution of complex hierarchical solutions, while rapid state transitions potentially allow the system to adapt to sudden environmental changes with minimal loss of functionality \cite{Levin}. 

A key question for designers of Digital Ecosystems is how the stability and diversity properties of biological ecosystems map to performance measures in digital systems. For a Digital Ecosystem the ultimate performance measure is user satisfaction, a system-specific property. However, assuming the motivation for engineering a Digital Ecosystem is the development of scalable, adaptive solutions to complex dynamic problems, certain generalisations can be made. Sustained diversity \cite{folke}, is a key requirement for dynamic adaptation. In Digital Ecosystems, diversity must be balanced against adaptive efficiency because maintaining large numbers of poorly-adapted solutions is costly. The exact form of this tradeoff will be guided by the specific requirements of the system in question. Stability \cite{Levin}, is likewise, a trade-off: we want the system to respond to environmental change with rapid adaptation, but not to be so responsive that mass extinctions deplete diversity or sudden state changes prevent control.

\section{The Digital Ecosystem}

We are concerned with the digital counterpart of biological ecosystems. However, the term \emph{digital ecosystem} has been used to describe a variety of concepts, which it now makes sense to review. Some of these refer to the existing networking infrastructure of the internet \cite{debook2, fiorina, XIMBIOTIX}, while several companies offer a \emph{digital ecosystem} service or solution, which involves enabling customers to use existing e-business solutions \cite{accenture, syntel, xewow}. The term is also being increasingly linked, yet undefined, to the future developments of \ac{ICT} adoption for e-business and e-commerce, to create so called \emph{business ecosystems} \cite{iansiti2004kan, nachira, papazoglou2001aot}. However, perhaps the most frequent references to \emph{digital ecosystems} arise in Artificial Life research, where they are created primarily to investigate aspects of biological and other complex systems \cite{sorakugun1995eas, grand1998ces, deAI1}. The extent to which these disparate systems resemble biological ecosystems varies, and frequently the word \emph{ecosystem} is merely used for branding purposes without any inherent ecological properties.

We consider Digital Ecosystems \cite{javaOne, bionetics, eveNet, eveSim} to be software systems that exploit the properties of biological ecosystems, which are robust, scalable, and self-organising \cite{Levin}. So, Digital Ecosystems provide a two-level optimisation scheme inspired by natural ecosystems, in which a decentralised peer-to-peer network forms an underlying tier of distributed agents. These agents then feed a second optimisation level based on an evolutionary algorithm that operates locally on single habitats (peers), aiming to find solutions that satisfy locally relevant constraints. The local search is sped up through this twofold process, providing better local optima as the distributed optimisation provides prior sampling of the search space by making use of computations already performed in other peers with similar constraints \cite{javaOne, bionetics}. The agents consist of an \emph{executable component} and an \emph{ontological description} \cite{wooldridge}. So, the Digital Ecosystem can be considered a \ac{MAS} \cite{wooldridge} which uses \emph{distributed evolutionary computing} \cite{cantupaz1998spg, stender1993pga} to combine suitable agents in order to meet user requests for applications.

The motivation for using parallel or distributed evolutionary algorithms is twofold. First, improving the speed of evolutionary processes by conducting concurrent evaluations of individuals in a population. Second, improving the problem-solving process by overcoming difficulties that face traditional evolutionary algorithms, such as maintaining diversity to avoid premature convergence \cite{muhlenbein1991eta, stender1993pga}. The fact that evolutionary computing manipulates a population of independent solutions actually makes it well suited for parallel computation architectures \cite{cantupaz1998spg}. There are several variants of distributed evolutionary computing, leading some to propose a taxonomy for their classification \cite{nowostawski1999pga}, with there being two main forms \cite{cantupaz1998spg, stender1993pga}: multiple-population/coarse-grained migration/island models \cite{lin1994cgp, cantupaz1998spg}, and single-population/fine-grained diffusion/neighbourhood models \cite{manderick1989fgp, stender1993pga}. Fine-grained \emph{diffusion} models \cite{manderick1989fgp, stender1993pga} assign one individual per processor. A local neighbourhood topology is assumed, and individuals are allowed to mate only within their neighbourhood, called a \emph{deme}. The demes overlap by an amount that depends on their shape and size, and in this way create an implicit migration mechanism. Each processor runs an identical evolutionary algorithm which selects parents from the local neighbourhood, produces an offspring, and decides whether to replace the current individual with an offspring. In the coarse-grained \emph{island} models \cite{lin1994cgp, cantupaz1998spg}, evolution occurs in multiple parallel sub-populations (islands), each running a local evolutionary algorithm, evolving independently with occasional \emph{migrations} of highly fit individuals among sub-populations. This model has also been used successfully in the determination of investment strategies in the commercial sector, in a product known as the Galapagos toolkit \cite{galapagos1, galapagos2}. However, all the \emph{islands} in this approach work on exactly the same problem, which makes it less analogous to biological ecosystems in which different locations can be environmentally different \cite{begon96}. 

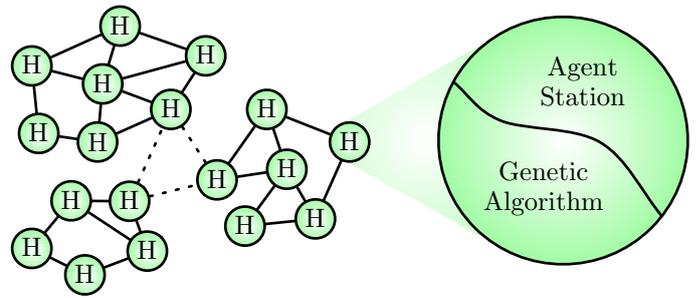
\tfigure{scale=1.0}{architecture2}{graffle}{Digital Ecosystem}{Optimisation architecture in which agents travel along the peer-to-peer connections; in every node (habitat) local optimisation is performed through an evolutionary algorithm, where the search space is determined by the agents present at the node.}{habnet}{-7mm}{}

The landscape, in energy-centric biological ecosystems, defines the connectivity between habitats \cite{begon96}.  Connectivity of nodes in the digital world is generally not defined by geography or spatial proximity, but by information or semantic proximity. For example, connectivity in a peer-to-peer network is based primarily on bandwidth and information content, and not geography. The island-models of \acl{DEC} use an information-centric model for the connectivity of nodes (\emph{islands}) \cite{lin1994cgp}. However, because it is generally defined for one-time use (to evolve a solution to one problem and then stop) it usually has a fixed connectivity between the nodes, and therefore a fixed topology \cite{cantupaz1998spg}. So, supporting evolution in the Digital Ecosystem, with a multi-objective \emph{selection pressure} (fitness landscape \cite{wright1932} with many peaks), requires a re-configurable network topology, such that habitat connectivity can be dynamically adapted based on the observed migration paths of the agents between the users within the habitat network. Based on the island-models of \acl{DEC} \cite{lin1994cgp}, each connection between the habitats is bi-directional and there is a probability associated with moving in either direction across the connection, with the connection probabilities affecting the rate of migration of the agents. However, additionally, the connection probabilities will be updated by the success or failure of agent migration using the concept of Hebbian learning \cite{hebb}: the habitats which do not successfully exchange agents will become less strongly connected, and the habitats which do successfully exchange agents will achieve stronger connections. This leads to a topology that adapts over time, resulting in a network that supports and resembles the connectivity of the user base. If we consider a \emph{business ecosystem}, network of \aclp{SME}, as an example user base; such business networks are typically small-world networks \cite{white2002nst, antionella}. They \setCap{many strongly connected clusters (communities), called \emph{sub-networks} (quasi-complete graphs), with a few connections between these clusters (communities) \cite{swn1}. Graphs with this topology have a very high clustering coefficient and small characteristic path lengths \cite{swn1}.}{archComTop} So, the Digital Ecosystem will take on a topology similar to that of the user base.

The novelty of our approach comes from the evolving populations being created in response to \emph{similar} requests. So whereas in the island-models of \acl{DEC} there are multiple evolving populations in response to one request \cite{lin1994cgp}, here there are multiple evolving populations in response to \emph{similar} requests. In our Digital Ecosystems different \setCap{requests are evaluated on separate \emph{islands} (populations), and so adaptation is accelerated by the sharing of solutions between evolving populations (islands), because they are working to solve similar requests (problems).}{similarCap}

The users \setCap{will formulate queries to the Digital Ecosystem by creating a request as a \emph{semantic description}, like those being used and developed in \acp{SOA} \cite{SOAsemantic}, specifying an application they desire and submitting it to their local peer (habitat).}{picUser} This description defines a metric for evaluating the \emph{fitness} of a composition of agents, as a distance function between the \emph{semantic description} of the request and the agents' \emph{ontological descriptions}. \setCap{A population is then instantiated in the user's habitat in response to the user's request, seeded from the agents available at their habitat.}{picUserReq} This allows the evolutionary optimisation to be accelerated in the following three ways: first, the habitat network provides a subset of the agents available globally, which is localised to the specific user it represents; second, making use of agent-sequences previously evolved in response to the user's earlier requests; and third, taking advantage of relevant agent-sequences evolved elsewhere in response to similar requests by other users. The population then proceeds to evolve the optimal agent-sequence(s) that fulfils the user request, and as the agents are the base unit for evolution, it searches the available agent-sequence combination space. For an evolved agent-sequence that is executed (instantiated) by the user, it then migrates to other peers (habitats) becoming hosted where it is useful, to combine with other agents in other populations to assist in responding to other user requests for applications.

\section{Simulation and Results}

We simulated the Digital Ecosystem, based upon our \acl{EOA}, and recorded key variables to determine whether it displayed behaviour typical of biological ecosystems. We created a simulation, following the \acl{EOA} from the previous section, using the Business Ecosystem of \aclp{SME} from \aclp{DBE} \cite{dbebkintro} as an example user base. 

Throughout the simulations we assumed a hundred users, which meant that at any time the number of users joining the network equalled those leaving. The habitats of the users were randomly connected at the start, to simulate the users going online for the first time. The users then produced agents (services) and requests for business applications. Initially, the users each deployed five agents to their habitats, for migration (distribution) to any habitats connected to theirs (i.e. their community within the Business Ecosystem). Users were simulated to deploy a new agent after the submission of three requests for business applications, and were chosen at random to submit their requests. \setCap{A simulated user request consisted of an abstract \emph{semantic description}, as a list of sets of numeric tuples to represent the properties of a desired business application}{semanticRequest}. The use of the \emph{numeric tuples} made it comparable to the \emph{semantic descriptions} of the services represented by the agents; while the \emph{list of sets} (two level hierarchy) and a much longer length provided sufficient complexity to support the sophistication of business applications.

The user requests were handled by the habitats instantiating evolving populations, which used evolutionary computing to find the optimal solution(s), agent-sequence(s). It was assumed that the users made their requests for business applications \emph{accurately}, and always used the response (agent-sequence) provided.

Populations of agents, $[A_1, A_1, A_2, ...]$, were evolved to solve user requests, seeded with agents and agent-sequences from the \emph{agent-pool} of the habitats in which they were instantiated. A dynamic population size was used to ensure exploration of the available combinatorial search space, which increased with the average length of the population's agent-sequences. The optimal combination of agents (agent-sequence) was evolved to the user request, $R$, by an artificial \emph{selection pressure} created by a \emph{fitness function} generated from the user request, $R$. An individual (agent-sequence) of the population consisted of a set of attributes, ${a_1, a_2, ...}$, and a user request essentially consisted of a set of required attributes, ${r_1, r_2, ...}$. So, the \emph{fitness function} for evaluating an individual agent-sequence, $A$, relative to a user request, $R$, was,
\vspace{6mm}
\begin{equation}
fitness(A,R) = \frac{1}{1 + \sum_{r \in R}{|r-a|}},
\label{ff}
\vspace{6mm}
\end{equation}
where $a$ is the member of $A$ such that the difference to the required attribute $r$ was minimised. Equation \ref{ff} was used to assign \emph{fitness} values between 0.0 and 1.0 to each individual of the current generation of the population, directly affecting their ability to replicate into the next generation. The evolutionary computing process was encoded with a low mutation rate, a fixed selection pressure and a non-trapping fitness function (i.e. did not get trapped at local optima). The type of selection used \emph{fitness-proportional} and \emph{non-elitist}. \emph{Fitness-proportional meaning that the \emph{fitter} the individual the higher its probability of} surviving to the next generation \cite{blickle1996css}. \emph{Non-elitist} meaning that the best individual from one generation was not guaranteed to survive to the next generation; it had a high probability of surviving into the next generation, but it was not guaranteed as it might have been mutated, \cite{eiben2003iec}. \emph{Crossover} (recombination) was then applied to a randomly chosen 10\% of the surviving population, a \emph{one-point crossover}, by aligning two parent individuals and picking a random point along their length, and at that point exchanging their tails to create two offspring \cite{eiben2003iec}. \emph{Mutations} were then applied to a randomly chosen 10\% of the surviving population; \emph{point mutations} were randomly located, consisting of \emph{insertions} (an agent was inserted into an agent-sequence), \emph{replacements} (an agent was replaced in an agent-sequence), and \emph{deletions} (an agent was deleted from an agent-sequence) \cite{lawrence1989hsd}. The issue of bloat was controlled by augmenting the \emph{fitness function} with a \emph{parsimony pressure \cite{soule1998ecg} which biased the search} to shorter agent-sequences, evaluating longer than average length agent-sequences with a reduced \emph{fitness}, and thereby providing a dynamic control limit which adapted to the average length of the ever-changing evolving agent populations.

\subsection{Ecological Succession}
\label{secSuccession}

We then compared some of the Digital Ecosystem's dynamics with those of biological ecosystems, to determine if it had been imbibed with the properties of biological ecosystems. A biological ecosystem develops from a simpler to a more mature state, by a process of \emph{succession}, where the genetic variation of the populations changes with time \cite{begon96}. \setCap{So, it becomes increasingly more complex through this process of succession, driven by the evolution of the populations within the ecosystem \cite{connell111msn}.}{succession} Equivalently, the Digital Ecosystem's increasing complexity comes from the agent populations being evolved to meet the dynamic selection pressures created by the user requests. 

\tfigure{width=3.25in}{successionNew}{pdf}{Ecological Succession}{\getCap{succession2} \getCap{succession3}. \getCap{succession}}{succession}{-3mm}{!h}{-5mm}

\setCap{The formation of a mature ecosystem}{succession2}, ecological succession, \setCap{is the slow, predictable, and orderly changes in the composition and structure of an ecological community, for which there are defined stages in the increasing complexity \cite{begon96}, as shown}{succession3} in Figure \ref{succession}. Succession may be initiated either by the formation of a new, unoccupied habitat (e.g., a lava flow or a severe landslide) or by some form of disturbance (e.g. fire, logging) of an existing community. The former case is often called \emph{primary succession}, and the latter \emph{secondary succession} \cite{begon96}. The trajectory of ecological change can be influenced by site conditions, by the interactions of the species present, and by more stochastic factors such as availability of colonists or seeds, or weather conditions at the time of disturbance. Some of these factors contribute to predictability of successional dynamics; others add more probabilistic elements \cite{gotelli1995pe}. Trends in ecosystem and community properties of succession have been suggested, but few appear to be general. For example, species diversity almost necessarily increases during early succession upon the arrival of new species, but may decline in later succession as competition eliminates opportunistic species and leads to dominance by locally superior competitors \cite{connell111msn}. Net Primary Productivity\footnote{Net Primary Productivity (NPP) is defined as the net flux of carbon from the atmosphere into green plants per unit time \cite{lawrence1989hsd}.}, biomass, and trophic level properties all show variable patterns over succession, depending on the particular system and site \cite{gotelli1995pe}. Generally, communities in early succession will be dominated by fast-growing, well-dispersed species, but as the succession proceeds these species will tend to be replaced by more competitive species \cite{begon96}.

We then considered existing theories of complexity for ecological succession and how it would apply to Digital Ecosystems, seeking a high-level understanding that would apply equally to both biological and digital ecosystems. As succession leads communities, of an ecosystem, to states of dynamic equilibrium within the environment \cite{begon96}, the complexity has to increase initially or there would not be an ecosystem, and presumably this increase eventually stops, because there must be a limit to how many species can be supported. The period in between is more complicated. If we consider the neutral biodiversity theory \cite{Hubbell}, which basically states network aspects of ecosystems are negligible, we would probably get a relatively smooth progression, because although you would get occasional extinctions, they would be randomly isolated events whose frequency would eventually balance arrivals, not self-organised crashes like in systems theory. In systems theory \cite{systemsTheory}, when a new species arrives in an ecological network, it can create a positive feedback loop that destabilises part of the network and drives some species to extinction. Ecosystems are constantly being perturbed, so it is reasonable to assume that a species that persists will probably be involved in a stabilising interaction with other species. So, the whole ecological network evolves to resist invasion. That would lead to a spiky succession process, perhaps getting less spiky over time. 

So, which theory is more applicable to the Digital Ecosystem depends on the extent that a species in the ecosystem acts independently, competing entities (smooth succession) \cite{Hubbell} versus tightly co-adapted ecological partners (spiky succession) \cite{systemsTheory}. Our Digital Ecosystem despite its relative complexity, is quite simplistic compared to biological ecosystems. It has the essential and fundamental processes, but no sophisticated social mechanisms. Therefore, the smooth succession of the neutral biodiversity theory \cite{Hubbell} is more probable.

As the Digital Ecosystem's increasing complexity comes from the Agent populations being evolved to meet the user requests, the effectiveness of these responses (agent-sequences) to user requests is the best available estimate of the Digital Ecosystem's complexity. So we measured the responses in the simulation of the Digital Ecosystem over a thousand user requests, i.e. until it had reached a mature state like a biological ecosystem \cite{begon96}, and graphed a typical run in Figure \ref{DigEcoSuc}. The range and diversity of agents at initial deployment were such that 70\% fulfilment of user requests was possible, increasing to 100\% fulfilment as more agents were deployed. 

\tfigure{scale=1.0}{DigEcoSuc}{graph}{Graph of Succession in the Digital Ecosystem}{\getCap{DigEcoSuc} Still, at \getCap{DigEcoSuc2}}{DigEcoSuc}{-7mm}{}{}

\label{ecosucexp}

The Digital Ecosystem performed as expected, adapting and improving over time, reaching a mature state as seen in the graph of Figure \ref{DigEcoSuc}. The succession of the Digital Ecosystem followed the smooth succession of the neutral biodiversity theory \cite{Hubbell}, shown by the \emph{tight} distribution, and equal density, of the points around the best fit curve of the graph in Figure \ref{DigEcoSuc}. At \setCap{the end of the simulation run, the agent-sequences had evolved and migrated over only ten generations, on average, and collectively had already reached near 70\% effectiveness for the user requests.}{DigEcoSuc2} \setCap{The formation of a mature biological ecosystem, ecological succession, is a relatively slow process \cite{begon96}, and the simulated Digital Ecosystem acted similarly in reaching a mature state.}{DigEcoSuc}

\subsection{Species Abundance}

We then considered \emph{relative abundance}, which \setCap{is the proportion of all organisms, in a community, belonging to a particular species \cite{Bell}. Relative abundance distributions provide a measure of inequalities in population size within an ecosystem, and in most biological ecosystems this distribution takes a log-normal form \cite{Bell}.}{speciesAbundance}

\tfigure{scale=1.0}{SpeciesHistogram}{graph}{Graph of Relative Abundance in the Digital Ecosystem}{Relative abundance \getCap{speciesAbundance} However, \getCap{specAbund2}.}{rad}{-7mm}{!h}{}

A snapshot of the agents (organisms) within the Digital Ecosystem, for a typical simulation run, was taken after a thousand user requests, i.e. once it had reached a mature state. In biology a species is a series of populations within which significant gene flow can and does occur, so groups of organisms showing a very similar genetic makeup \cite{lawrence1989hsd}. We therefore chose to define species within Digital Ecosystems similarly, as a grouping of genetically similar digital organisms (based on their semantic descriptions), with no more than 10\% variation within the species group. Relative abundance was calculated for each species and grouped by frequency in Figure \ref{rad}. In contrast to expectations from biological ecosystems, relative abundance in \setCap{the Digital Ecosystem did not conform to the expected log-normal}{specAbund2} \cite{Bell}. We speculate that the high frequency for the lowest relative abundance was caused by the dynamically re-configurable topology of the habitat network, which allowed species of small abundance to survive as their respective habitats were clustered by the Digital Ecosystem. It is also quite possible that this effect also skewed other frequencies for the relative abundance measure.

\subsection{Species-Area Relationship}

\setCap{The \emph{species-area} relationship measures diversity relative to spatial scale \cite{sizling2004pls}. In the Digital Ecosystem, this relationship represents how similar solutions are to one another at different habitat scales. The species-area relationship is commonly found to follow a power law in biological ecosystems}{speciesArea} \cite{sizling2004pls}.

\tfigure{scale=1.0}{speciesArea}{graph}{Graph of Species-Area in the Digital Ecosystem}{\getCap{speciesArea}, which the Digital Ecosystem also demonstrates.}{sar}{-7mm}{!h}{}

Again, a snapshot of the agents (organisms) within the Digital Ecosystem, for a typical simulation run, was taken once it had reached a mature state, after a thousand user r equests. For this experiment, we assumed each habitat to have an area of one unit. Then, the number of species, at $n$ randomly chosen habitats, was measured, where $n$ ranged between one and a hundred. For each $n$, ten sets of measurements were taken at different random sets of habitats to calculate averaged results, and the $log_{10}$ values of these results are depicted in the graph of Figure \ref{sar}. The distribution of species diversity over a spatial scale in the Digital Ecosystem demonstrates behaviour similar to biological ecosystems, also following a power law \cite{sizling2004pls}. However, diversity at fine spatial scales appears to be lower than predicted by the line of best fit. This may be explained by higher specialisation at some habitats, making them more like micro-habitats in terms of a reduced species diversity \cite{lawrence1989hsd}.

\section{Conclusion}

In simulation, we compared the Digital Ecosystem's dynamics to those of biological ecosystems. The \emph{ecological succession}, measured by the responsiveness to user requests, conformed to expectations from biological ecosystems \cite{Hubbell}: improving over time, before approaching a plateau. As the evolutionary self-organisation of an ecosystem is a slow process, even the accelerated form present in Digital Ecosystems, it reached only 70\% responsiveness, showing \emph{potential} for improvement. In the \emph{species abundance} experiment the Digital Ecosystem did not conform to the log-normal distribution usually found in biological ecosystems \cite{Bell}. The high frequency for the lowest relative abundance was probably caused by the dynamically re-configurable topology of the habitat network, which allowed species of small abundance to survive as their habitats were clustered by the Digital Ecosystem. In the \emph{species-area} experiment, which measures diversity relative to spatial scale, the Digital Ecosystem did follow the power law commonly found in biological ecosystems \cite{sizling2004pls}. The \emph{species diversity} at fine spatial scales was lower than predicted by the line of best fit, and may be explained by the high specialisation at some habitats, making them more like micro-habitats, including a reduced \emph{species diversity} \cite{lawrence1989hsd}. The \emph{majority} of the experimental results indicate that Digital Ecosystems behave like their biological counterparts, and suggest that incorporating ideas from theoretical ecology can contribute to useful self-organising properties in Digital Ecosystems, which can assist in generating scalable solutions to complex dynamic problems.

By comparing and contrasting the relevant theoretical ecology, with the anticipated requirements of Digital Ecosystems, we examined how ecological features may emerge in some systems designed for adaptive problem solving. Specifically, we suggested that Digital Ecosystems, like a biological ecosystem, will usually consist of self-replicating agents that interact both with one another and with an external environment \cite{begon96}. Population dynamics and evolution, spatial and network interactions, and complex dynamic fitness landscapes, will all influence the behaviour of these systems. Many of these properties can be understood via well-known ecological models \cite{MacArthur, Hubbell}, with a further body of theory that treats ecosystems as \acl{CAS} \cite{Levin}. These models provide a theoretical basis for the occurrence of self-organisation, in digital and biological ecosystems, resulting from the interactions among the agents and their environment, leading to complex non-linear behaviour \cite{MacArthur, Hubbell, Levin}; and it is this property that provides the underlying potential for scalable problem-solving in digital environments.

Creating the digital counterpart of biological ecosystems was not without apparent compromises; the information-centric dynamically re-configurable network topology, and the \emph{species abundance} result inconsistent with biological ecosystems. The Digital Ecosystem requires a re-configurable network topology, to support the constantly changing multi-objective information-centric \emph{selection pressures} of the user base. Hence, using the concept of Hebbian learning \cite{hebb}, habitat connectivity is dynamically adapted based on the observed migration paths of the agents within the habitat network. The dynamically re-configurable network topology probably caused the Digital Ecosystem not to conform, in the \emph{species abundance} experiment, to the log-normal distribution expected from biological ecosystems \cite{Bell}. We would argue that these differences are not compromises, but features unique to Digital Ecosystems. As we discussed earlier, biomimicry, when done well, is not slavish imitation; it is inspiration using the principles which nature has demonstrated to be successful design strategies \cite{biomimicry}. Hypothetically, if there were an abstract definition of an ecosystem, defined as an abstract \emph{ecosystem} class, \setCap{then the \emph{Digital Ecosystem} and \emph{biological ecosystem} classes would both inherit from the abstract \emph{ecosystem} class, but implement its attributes differently}{ecoClass}. \setCap{So, we would argue that the apparent compromises in mimicking biological ecosystems are actually features unique to Digital Ecosystems.}{eco2Class}

Service-oriented architectures promise to provide potentially huge numbers of services that programmers can combine via standardised interfaces, to create increasingly sophisticated and distributed applications \cite{SOApaper2}. The Digital Ecosystem extends this concept with the automatic combining of available and applicable services in a scalable architecture to meet user requests for applications. This is made possible by a fundamental paradigm shift, from a \emph{pull}-oriented approach to a \emph{push}-oriented approach. So, instead of the \emph{pull}-oriented approach of generating applications only upon request in \aclp{SOA} \cite{singh2005soc}, the Digital Ecosystem follows a \emph{push}-oriented approach of distributing and composing applications pre-emptively, as well as upon request. Although the use of \aclp{SOA} in the definition of Digital Ecosystems provides a predisposition to business \cite{krafzig2004ess}, it does not preclude other more general uses. The \acl{EOA} definition of Digital Ecosystems is intended to be inclusive and interoperable with other technologies, in the same way that the definition of \aclp{SOA} is with \emph{grid computing and other} technologies \cite{singh2005soc}. For example, habitats could be executed using a distributed processing arrangement, such as \emph{cloud computing} \cite{weiss2007cc}, which would be possible because the habitat network topology is information-centric (instead of location-centric).

We have confirmed the fundamentals for a new class of system, Digital Ecosystems, created through combining understanding from theoretical ecology, evolutionary theory, \aclp{MAS}, \acl{DEC}, and \aclp{SOA}. Digital Ecosystems, where the word \emph{ecosystem} is more than just a metaphor, being the digital counterpart of biological ecosystems, and therefore having their desirable properties, such as scalability and self-organisation. It is a complex system that shows emergent behaviour, being more than the sum of its constituent parts.

\section*{Acknowledgments}

We thank P. Dini and P. De Wilde for constructive comments. This work was supported by the European Commission under the EU project Digital Business Ecosystems (contract number 507953 \cite{DBE})

\bibliographystyle{IEEEtran.bst}
\bibliography{../../../PhDthesis/references}

\begin{thebibliography}{10}
\providecommand{\url}[1]{#1}
\csname url@rmstyle\endcsname
\providecommand{\newblock}{\relax}
\providecommand{\bibinfo}[2]{#2}
\providecommand\BIBentrySTDinterwordspacing{\spaceskip=0pt\relax}
\providecommand\BIBentryALTinterwordstretchfactor{4}
\providecommand\BIBentryALTinterwordspacing{\spaceskip=\fontdimen2\font plus
\BIBentryALTinterwordstretchfactor\fontdimen3\font minus
  \fontdimen4\font\relax}
\providecommand\BIBforeignlanguage[2]{{%
\expandafter\ifx\csname l@#1\endcsname\relax
\typeout{** WARNING: IEEEtran.bst: No hyphenation pattern has been}%
\typeout{** loaded for the language `#1'. Using the pattern for}%
\typeout{** the default language instead.}%
\else
\language=\csname l@#1\endcsname
\fi
#2}}

\bibitem{newsArticle1}
S.~Miller, ``Aspect-oriented programming takes aim at software complexity,''
  \emph{Computer}, vol.~34, pp. 18--21, 2001.

\bibitem{slashdot}
H.~Sutter, ``The free lunch is over: A fundamental turn toward concurrency in
  software,'' \emph{Dr. Dobb's Journal}, 2005, cr newsArticle2.

\bibitem{newsArticle3}
\BIBentryALTinterwordspacing
J.~Markoff, ``Faster chips are leaving programmers in their dust,'' New York
  Times, Tech. Rep., 2007. [Online]. Available:
  \url{http://www.nytimes.com/2007/12/17/technology/17chip.html}
\BIBentrySTDinterwordspacing

\bibitem{lyytinen2001nwn}
K.~Lyytinen and Y.~Yoo, ``The next wave of nomadic computing: A research agenda
  for information systems research,'' \emph{Sprouts: Working Papers on
  Information Systems}, vol.~1, pp. 1--20, 2001.

\bibitem{reef3}
S.~McIlraith, C.~Son, and H.~Zeng, ``Semantic web services,'' \emph{IEEE
  Intelligent Systems}, vol.~16, pp. 46--53, 2001.

\bibitem{reef5}
S.~Narayanan and S.~McIlraith, ``Simulation, verification and automated
  composition of web services,'' in \emph{international conference on World
  Wide Web}.\hskip 1em plus 0.5em minus 0.4em\relax ACM Press, 2002, pp.
  77--88.

\bibitem{reef1}
N.~Milanovic and M.~Malek, ``Current solutions for web service composition,''
  \emph{IEEE Internet Computing}, vol.~8, pp. 51--59, 2004.

\bibitem{reef4}
J.~Rao and X.~Su, ``A survey of automated web service composition methods,'' in
  \emph{Semantic Web Services and Web Process Composition}, J.~Cardoso and
  A.~Sheth, Eds.\hskip 1em plus 0.5em minus 0.4em\relax Springer, 2004, pp.
  43--54.

\bibitem{curbera2002uws}
F.~Curbera, M.~Duftler, R.~Khalaf, W.~Nagy, N.~Mukhi, and S.~Weerawarana,
  ``Unraveling the web services web: An introduction to {SOAP}, {WSDL}, and
  {UDDI},'' \emph{IEEE Internet Computing}, vol.~6, pp. 86--93, 2002.

\bibitem{SOAstandards}
\BIBentryALTinterwordspacing
B.~Violino. (2007) How to navigate a sea of {SOA} standards. [Online].
  Available:
  \url{http://www.cio.com/article/104007/How_to_Navigate_a_Sea_of_SOA_Standard%
s}
\BIBentrySTDinterwordspacing

\bibitem{biomimicry}
J.~Benyus, \emph{Biomimicry, Innovation Inspired by Nature}.\hskip 1em plus
  0.5em minus 0.4em\relax Harper Collins Publishers, 2002.

\bibitem{Levin}
S.~Levin, ``Ecosystems and the biosphere as complex adaptive systems,''
  \emph{Ecosystems}, vol.~1, pp. 431--436, 1998.

\bibitem{eiben2003iec}
A.~Eiben and J.~Smith, \emph{Introduction to Evolutionary Computing}.\hskip 1em
  plus 0.5em minus 0.4em\relax Springer, 2003.

\bibitem{debook2}
P.~Denning and R.~Metcalfe, \emph{Beyond Calculation: The Next Fifty Years of
  Computing}.\hskip 1em plus 0.5em minus 0.4em\relax Springer, 1997.

\bibitem{fiorina}
\BIBentryALTinterwordspacing
C.~Fiorina. (2000) The digital ecosystem. [Online]. Available:
  \url{http://www.hp.com/hpinfo/execteam/speeches/fiorina/ceo_worldres_00.html}
\BIBentrySTDinterwordspacing

\bibitem{XIMBIOTIX}
\BIBentryALTinterwordspacing
Ximbiotix. (2005) About the digital ecosystem. [Online]. Available:
  \url{http://www.ximbiotix.com/desktop/the-digital-ecosystem/about-the-digita%
l-ecosystem.cfm}
\BIBentrySTDinterwordspacing

\bibitem{accenture}
\BIBentryALTinterwordspacing
D.~Bennett, ``Digital transformation in the entertainment industry - embracing
  the fully digital ecosystem,'' Accenture, Tech. Rep., 2006. [Online].
  Available:
  \url{http://www.accenture.com/NR/rdonlyres/A58111E4-22E5-4DDD-B3DE-FB3741F00%
52F/0/EmbracingDigitalEco.pdf}
\BIBentrySTDinterwordspacing

\bibitem{syntel}
\BIBentryALTinterwordspacing
M.~Kulkarni and R.~Kreutzer, ``Building your own digital ecosystem{:} a
  holistic approach to enterprise integration,'' Syntel, Tech. Rep., 2006.
  [Online]. Available:
  \url{http://www.syntelinc.com/uploadedFiles/Syntel_DigitalEcosystem.pdf}
\BIBentrySTDinterwordspacing

\bibitem{xewow}
\BIBentryALTinterwordspacing
M.~Vandenberghe, ``Digital ecosystem solution - the business factory,'' XeWOW,
  Tech. Rep., 2006. [Online]. Available:
  \url{http://themaddesigner.free.fr/XeWOW%20White%20Paper.pdf}
\BIBentrySTDinterwordspacing

\bibitem{iansiti2004kan}
M.~Iansiti and R.~Levien, \emph{The Keystone Advantage: What the New Dynamics
  of Business Ecosystems Mean for Strategy, Innovation, and
  Sustainability}.\hskip 1em plus 0.5em minus 0.4em\relax Harvard Business
  School Press, 2004.

\bibitem{nachira}
\BIBentryALTinterwordspacing
F.~Nachira, ``Towards a network of digital business ecosystems fostering the
  local development,'' Directorate General Information Society and Media,
  European Commission, Tech. Rep., 2002. [Online]. Available:
  \url{http://www.digital-ecosystems.org/doc/discussionpaper.pdf}
\BIBentrySTDinterwordspacing

\bibitem{papazoglou2001aot}
M.~Papazoglou, ``Agent-oriented technology in support of e-business,''
  \emph{Communications of the ACM}, vol.~44, pp. 71--77, 2001.

\bibitem{sorakugun1995eas}
K.~Soraku-gun, ``An evolutionary approach to synthetic biology: Zen and the art
  of creating life,'' in \emph{Artificial Life: An Overview}, C.~Langton,
  Ed.\hskip 1em plus 0.5em minus 0.4em\relax MIT Press, 1995, pp. 195--226.

\bibitem{grand1998ces}
S.~Grand and D.~Cliff, ``Creatures: Entertainment software agents with
  artificial life,'' \emph{Autonomous Agents and Multi-Agent Systems}, vol.~1,
  pp. 39--57, 1998.

\bibitem{deAI1}
D.~Cliff and S.~Grand, ``The creatures global digital ecosystem,''
  \emph{Artificial Life}, vol.~5, pp. 77--93, 1999.

\bibitem{begon96}
M.~Begon, J.~Harper, and C.~Townsend, \emph{Ecology: Individuals, Populations
  and Communities}.\hskip 1em plus 0.5em minus 0.4em\relax Blackwell
  Publishing, 1996.

\bibitem{goldberg}
D.~Goldberg, \emph{Genetic algorithms in search, optimization, and machine
  learning}.\hskip 1em plus 0.5em minus 0.4em\relax Addison-Wesley, 1989, cr
  ec21 goldberg89.

\bibitem{wright1932}
S.~Wright, ``The roles of mutation, inbreeding, crossbreeding and selection in
  evolution,'' in \emph{International Congress on Genetics}, D.~Jones,
  Ed.\hskip 1em plus 0.5em minus 0.4em\relax Brooklyn botanic garden, 1932, pp.
  356--366.

\bibitem{morrison2004dea}
R.~Morrison, \emph{Designing Evolutionary Algorithms For Dynamic
  Environments}.\hskip 1em plus 0.5em minus 0.4em\relax Springer, 2004.

\bibitem{chmbers2001phg}
L.~Chmbers, \emph{The practical handbook of genetic algorithms:
  applications}.\hskip 1em plus 0.5em minus 0.4em\relax CRC Press, 2001.

\bibitem{levins1969sda}
R.~Levins, ``Some demographic and genetic consequences of environmental
  heterogeneity for biological control,'' \emph{Bulletin of the Entomological
  Society of America}, vol.~15, pp. 237--240, 1969.

\bibitem{hanski1999me}
I.~Hanski, \emph{Metapopulation Ecology}.\hskip 1em plus 0.5em minus
  0.4em\relax Oxford University Press, 1999.

\bibitem{hanski2003mtf}
I.~Hanski and O.~Ovaskainen, ``Metapopulation theory for fragmented
  landscapes,'' \emph{Theoretical Population Biology}, vol.~64, pp. 119--127,
  2003.

\bibitem{Greenetal2006}
D.~Green, N.~Klomp, G.~Rimmington, and S.~Sadedin, \emph{Complexity in
  Landscape Ecology}.\hskip 1em plus 0.5em minus 0.4em\relax Springer, 2006.

\bibitem{suzie}
D.~Green and S.~Sadedin, ``Interactions matter- complexity in landscapes and
  ecosystems,'' \emph{Ecological Complexity}, vol.~2, pp. 117--130, 2005, cr
  GreenSadedin.

\bibitem{Bell}
G.~Bell, ``The distribution of abundance in neutral communities,''
  \emph{American Naturalist}, vol. 396, pp. 606--617, 2000.

\bibitem{sizling2004pls}
A.~Sizling and D.~Storch, ``Power-law species-area relationships and
  self-similar species distributions within finite areas,'' \emph{Ecology
  Letters}, vol.~7, pp. 60--68, 2004.

\bibitem{Hubbell}
S.~Hubbell, \emph{The Unified Neutral Theory of Biodiversity and
  Biogeography}.\hskip 1em plus 0.5em minus 0.4em\relax Princeton University
  Press, 2001.

\bibitem{Sole}
R.~Sol{\'e}, D.~Alonso, and A.~McKane, ``Self-organized instability in complex
  ecosystems,'' \emph{Philosophical Transactions: Biological Sciences}, vol.
  357, pp. 667--681, 2002.

\bibitem{GreenKirley}
D.~Green and M.~Kirley, ``Adaptation, diversity and spatial patterns,''
  \emph{International Journal of Knowledge-Based Intelligent Engineering
  Systems}, vol.~4, pp. 184--190, 2000.

\bibitem{Greenetal2000}
D.~Green, D.~Newth, and M.~Kirley, ``Connectivity and catastrophe - towards a
  general theory of evolution,'' in \emph{International Conference on
  Artificial Life}, M.~Bedau, Ed.\hskip 1em plus 0.5em minus 0.4em\relax MIT
  Press, 2000, pp. 153--161.

\bibitem{Greenetalinpress}
D.~Green, T.~Leishman, and S.~Sadedin, ``Dual phase evolution: a mechanism for
  self-organization in complex systems,'' in \emph{International Conference on
  Complex Systems}, A.~Minai, D.~Braha, and Y.~Bar-Yam, Eds.\hskip 1em plus
  0.5em minus 0.4em\relax Springer, 2006.

\bibitem{ducheyne2003fiu}
E.~Ducheyne, B.~De~Baets, and R.~De~Wulf, ``Is fitness inheritance useful for
  real-world applications,'' in \emph{Evolutionary Multi-criterion
  Optimization}, C.~Fonseca, Ed.\hskip 1em plus 0.5em minus 0.4em\relax
  Springer, 2003, pp. 31--42.

\bibitem{levin1999fdc}
S.~Levin, \emph{Fragile dominion: complexity and the commons}.\hskip 1em plus
  0.5em minus 0.4em\relax Perseus Books Group, 1999.

\bibitem{folke}
C.~Folke, S.~Carpenter, B.~Walker, M.~Scheffer, T.~Elmqvist, L.~Gunderson, and
  C.~Holling, ``Regime shifts, resilience, and biodiversity in ecosystem
  management,'' \emph{Annual Review of Ecology, Evolution, and Systematics},
  vol.~35, pp. 557--581, 2004.

\bibitem{newman1997mme}
M.~Newman, ``A model of mass extinction,'' \emph{Journal of Theoretical
  Biology}, vol. 189, pp. 235--252, 1997.

\bibitem{javaOne}
\BIBentryALTinterwordspacing
G.~Briscoe, M.~Chli, and M.~Vidal, ``{C}reating a {D}igital {E}cosystem:
  {S}ervice-oriented architectures with distributed evolutionary computing
  ({BOF}-0759),'' in \emph{JavaOne Conference}.\hskip 1em plus 0.5em minus
  0.4em\relax Sun Microsystems, 2006. [Online]. Available:
  \url{http://arxiv.org/0712.4159}
\BIBentrySTDinterwordspacing

\bibitem{bionetics}
G.~Briscoe, ``Digital {E}cosystems: Evolving service-oriented architectures,''
  in \emph{Conference on Bio Inspired mOdels of NETwork, Information and
  Computing Systems}.\hskip 1em plus 0.5em minus 0.4em\relax IEEE Press, 2006.

\bibitem{eveNet}
\BIBentryALTinterwordspacing
C.~Masuch. (2008) Evolutionary environment network. [Online]. Available:
  \url{http://evenet.sourceforge.net/}
\BIBentrySTDinterwordspacing

\bibitem{eveSim}
\BIBentryALTinterwordspacing
T.~Kurz. (2008) Ev{E} simulator. [Online]. Available:
  \url{http://evesim.sourceforge.net/}
\BIBentrySTDinterwordspacing

\bibitem{wooldridge}
M.~Wooldridge, \emph{Introduction to MultiAgent Systems}.\hskip 1em plus 0.5em
  minus 0.4em\relax Wiley, 2002.

\bibitem{cantupaz1998spg}
E.~Cantu-Paz, ``A survey of parallel genetic algorithms,'' \emph{R{\'e}seaux et
  syst{\`e}mes r{\'e}partis, Calculateurs Parall{\`e}les}, vol.~10, pp.
  141--171, 1998.

\bibitem{stender1993pga}
J.~Stender, \emph{Parallel Genetic Algorithms: Theory and Applications}.\hskip
  1em plus 0.5em minus 0.4em\relax IOS Press, 1993.

\bibitem{muhlenbein1991eta}
H.~Muhlenbein, ``Evolution in time and space - the parallel genetic
  algorithm,'' \emph{Foundations of Genetic Algorithms}, vol.~1, pp. 316--337,
  1991.

\bibitem{nowostawski1999pga}
M.~Nowostawski and R.~Poli, ``Parallel genetic algorithm taxonomy,'' in
  \emph{International Conference on Knowledge-Based Intelligent Information
  Engineering Systems}, L.~Jain, Ed.\hskip 1em plus 0.5em minus 0.4em\relax
  IEEE Press, 1999, pp. 88--92.

\bibitem{lin1994cgp}
S.~Lin, W.~Punch~III, and E.~Goodman, ``Coarse-grain parallel genetic
  algorithms: categorization and new approach,'' in \emph{Symposium on Parallel
  and Distributed Processing}.\hskip 1em plus 0.5em minus 0.4em\relax IEEE
  Press, 1994, pp. 28--37.

\bibitem{manderick1989fgp}
B.~Manderick and P.~Spiessens, ``Fine-grained parallel genetic algorithms,'' in
  \emph{International Conference on Genetic Algorithms}, J.~Schaffer, Ed.\hskip
  1em plus 0.5em minus 0.4em\relax Morgan Kaufmann Publishers, 1989, pp.
  428--433.

\bibitem{galapagos1}
\BIBentryALTinterwordspacing
M.~Ward. (2004) Life offers lessons for business. [Online]. Available:
  \url{http://news.bbc.co.uk/1/hi/technology/3752725.stm}
\BIBentrySTDinterwordspacing

\bibitem{galapagos2}
\BIBentryALTinterwordspacing
{Codefarm Software Limited}. (2008) Codefarm - technology for structured
  credit. [Online]. Available: \url{http://www.codefarm.com/}
\BIBentrySTDinterwordspacing

\bibitem{hebb}
D.~Hebb, \emph{The Organization of Behavior}.\hskip 1em plus 0.5em minus
  0.4em\relax Wiley, 1949.

\bibitem{white2002nst}
D.~White and M.~Houseman, ``The navigability of strong ties: Small worlds, tie
  strength, and network topology,'' \emph{Complexity}, vol.~8, pp. 72--81,
  2002.

\bibitem{antionella}
X.~Yang, ``Chaos in small-world networks,'' \emph{Physical Review E}, vol.~63,
  pp. 046\,206:1--4, 2001.

\bibitem{swn1}
D.~Watts and S.~Strogatz, ``Collective dynamics of `small-world' networks,''
  \emph{Nature}, vol. 393, pp. 440--442, 1998.

\bibitem{SOAsemantic}
P.~Rajasekaran, J.~Miller, K.~Verma, and A.~Sheth, ``Enhancing web services
  description and discovery to facilitate composition,'' in \emph{Semantic Web
  Services and Web Process Composition}, J.~Cardoso and A.~Sheth, Eds.\hskip
  1em plus 0.5em minus 0.4em\relax Springer, 2004, pp. 55--68.

\bibitem{dbebkintro}
F.~Nachira, P.~Dini, and A.~Nicolai, ``A network of digital business ecosystems
  for europe: Roots, processes and perspectives,'' in \emph{Digital {B}usiness
  {E}cosystems}, F.~Nachira, A.~Nicolai, P.~Dini, M.~Le~Louarn, and
  L.~Rivera~Le{\'o}n, Eds.\hskip 1em plus 0.5em minus 0.4em\relax European
  {C}ommission, 2007, pp. 1--20.

\bibitem{blickle1996css}
T.~Blickle and L.~Thiele, ``A comparison of selection schemes used in
  evolutionary algorithms,'' \emph{Evolutionary Computation}, vol.~4, pp.
  361--394, 1996.

\bibitem{lawrence1989hsd}
E.~Lawrence, \emph{Henderson's dictionary of biological terms}.\hskip 1em plus
  0.5em minus 0.4em\relax Pearson Education, 2005.

\bibitem{soule1998ecg}
T.~Soule and J.~Foster, ``Effects of code growth and parsimony pressure on
  populations in genetic programming,'' \emph{Evolutionary Computation},
  vol.~6, pp. 293--309, 1998.

\bibitem{connell111msn}
J.~Connell and R.~Slatyer, ``Mechanisms of succession in natural communities
  and their role in community stability and organization,'' \emph{The American
  Naturalist}, vol. 111, pp. 1119--1144, 1977.

\bibitem{gotelli1995pe}
N.~Gotelli, \emph{A primer of ecology}.\hskip 1em plus 0.5em minus 0.4em\relax
  Sinauer Associates, 1995.

\bibitem{systemsTheory}
R.~Ellen, \emph{Environment, Subsistence, and System: The Ecology of
  Small-Scale Social Formations}.\hskip 1em plus 0.5em minus 0.4em\relax
  Cambridge University Press, 1982.

\bibitem{MacArthur}
R.~MacArthur and E.~Wilson, \emph{The Theory of Island Biogeography}.\hskip 1em
  plus 0.5em minus 0.4em\relax Princeton University Press, 1967.

\bibitem{SOApaper2}
M.~Papazoglou and D.~Georgakopoulos, ``Service-oriented computing,''
  \emph{Communications of the ACM}, vol.~46, pp. 25--28, 2003, cr 4.5.

\bibitem{singh2005soc}
M.~Singh and M.~Huhns, \emph{Service-Oriented Computing: Semantics, Processes,
  Agents}.\hskip 1em plus 0.5em minus 0.4em\relax Wiley, 2005.

\bibitem{krafzig2004ess}
D.~Krafzig, K.~Banke, and D.~Slama, \emph{Enterprise SOA: Service-Oriented
  Architecture Best Practices}.\hskip 1em plus 0.5em minus 0.4em\relax Prentice
  Hall, 2004.

\bibitem{weiss2007cc}
A.~Weiss, ``Computing in the clouds,'' \emph{netWorker}, vol.~11, pp. 16--25,
  ACM Press, 2007.

\bibitem{DBE}
\BIBentryALTinterwordspacing
{EU Framework 6 Integrated Project}. Digital {B}usiness {E}cosystems. [Online].
  Available: \url{http://www.digital-ecosystem.org/DBE_Main/about}
\BIBentrySTDinterwordspacing

\end{thebibliography}

\end{document}